\documentclass[twoside,11pt]{article}

\usepackage{jmlr2e}
\usepackage{amsmath}
\usepackage{multicol}
\usepackage{subcaption}
\usepackage{bm}

\usepackage{color}
\newcommand{\myequation}{\begin{equation}}
\newcommand{\myendequation}{\end{equation}}
\let\[\myequation
\let\]\myendequation

\newcommand{\fracpartial}[2]{\frac{\partial #1}{\partial  #2}}

\ShortHeadings{Exploiting Transitivity for Top-k Selection with Score-Based Dueling Bandits}{Groves and Branke}
\firstpageno{1}

\begin{document}

\title{Exploiting Transitivity for Top-k Selection\\ with Score-Based Dueling Bandits}

\author{\name Matthew Groves \email M.J.Groves@warwick.ac.uk \\
       \addr Mathematics Institute\\
       University of Warwick\\
       Coventry, UK
       \AND
       \name Juergen Branke \email Juergen.Branke@wbs.ac.uk \\
       \addr Warwick Business School\\
       University of Warwick\\
       Coventry, UK}

\editor{}

\maketitle

\begin{abstract}%   <- trailing '%' for backward compatibility of .sty file
We consider the problem of top-k subset selection in Dueling Bandit problems with score information. Real-world pairwise ranking problems often exhibit a high degree of transitivity and prior work has suggested sampling methods that exploit such transitivity through the use of parametric preference models like the Bradley-Terry-Luce (BTL) and Thurstone models. To date, this work has focused on cases where sample outcomes are win/loss binary responses. We extend this to selection problems where sampling results contain quantitative information by proposing a Thurstonian style model and adapting the Pairwise Optimal Computing Budget Allocation for subset selection (POCBAm) sampling method to exploit this model for efficient sample selection. We compare the empirical performance against standard POCBAm and other competing algorithms.
\end{abstract}


\section{Introduction}
Multi-Armed Bandit (MAB) problems are an active area of research that consider the online learning of optimal decision alternatives from a given set, using information obtained from sampling. The standard MAB setting assumes that sampling information can be collected directly and independently for individual alternatives. However, direct measurement of alternatives can be impractical or impossible. For example, when ranking game players or sports teams, it is often very difficult to assess the ability of a player independently, but pairwise information about the player's performance in the form of game results against other players may be readily available.
%Dueling Bandit problems are an active area of research within the Multi-Armed Bandit (MAB) literature that consider the online learning of optimal decision alternatives from a given set, using information collected from pairs of alternatives. Whilst the standard MAB setting assumes that sampling information about individual alternatives can be collected independently, this assumption is often limiting as, for many applications, direct measurement of alternatives can be impractical or impossible. For example, when ranking game players or sports teams, it is often very difficult to assess the ability of a player independently, but pairwise information about the player's performance in the form of game results against other players may be readily available.
Well known methods for extracting player rankings from given sets of pairwise comparison outcomes like the ELO ranking \citep{elo} or TrueSkill ranking \citep{guo2012} have been widely applied to provide player rankings for a variety of games including Baseball, Chess, Go, and XBox\texttrademark gamers. 

Dueling Bandit approaches actively learn ratings by iteratively choosing pairs to compare, thereby allowing them to collect information more efficiently by selecting the most relevant or informative samples. These methods are applicable to search engine optimization or personalized advertising, where an advertiser may present two links or adverts to a user and receive feedback based on which link is clicked, or in evolutionary reinforcement learning, where the learner wishes to identify the high performing individuals in the evolutionary population and is free to choose which players to test against each other in simulations. Previous Dueling Bandit methods have generally focused on ranking problems where sample outcomes are qualitative, expressing solely a preference for one alternative in the pair over the other, or a simple win/loss game result. This is an important scenario as, especially in cases where collecting a pairwise sample involves human interaction, it is often simpler and more reliable to request a simple binary preference response. However, in some cases more information may be available from the comparison. For example, many pairwise games like Football, Tennis or Basketball provide natural numerical score information, which can provide a richer source of ranking information than simple win/loss results.

In this work, we focus on the case of selecting the top-k $\geq 1$ alternatives from an available set, where pairwise sampling results represent quantitative information about the degree of preference for one alternative over the other. Just as many preference-based dueling bandit methods often utilize parametric fitness models to encode reasonable assumptions on transitive relations between comparison outcomes and alternative rankings, our proposed method uses a similar approach, adapting the commonly used Thurstone model \citep{thurstone} for the quantiative case.
Our contributions can be summarized as follows:
\begin{itemize}
    \item We adapt a parametric preference model to cases where pairwise comparison outcomes are quantitative and continuous valued.
    \item We present a novel, adaptive dueling bandit method for pairwise top-k selection that can efficiently exploit this preference model whilst remaining robust to pairwise intransitivity.
    \item We demonstrate the performance of our method and compare against other published approaches on a variety of top-k selection tasks.
\end{itemize}
The paper is organised as follows: In the next section we present a description of the pairwise top-k selection problem, before giving an overview of related works in Section \ref{rw}. Section \ref{pocbam} defines the parametric ranking model and describes the proposed ML-POCBAm sampling method in detail. In Section \ref{tests} we compare the performance of ML-POCBAm against alternative methods on cases where the assumed preference model accurately describes the underlying alternative ranking. In Section \ref{nt}, we investigate the effect of noise and inconsistency in the underlying ranking, and propose a simple adaptation to ML-POCBAm to improve its robustness in such cases. Our final experiments in Section \ref{poker_experiments} apply ML-POCBAm to the top-k selection of artificial Texas Hold'em players.

\section{Problem Description}\label{pd}
Given a population $\mathcal{P} = \{a_{1},...,a_{K}\}$ of $K\geq2$ alternatives, we consider the problem of identifying the subset of the $k \geq 1$ from $\mathcal{P}$ highest ranked alternatives. We assume that fitness of alternatives cannot be measured directly, and is instead estimated relative to their peers in $\mathcal{P}$, determined by a weighted average of performance in pairwise tests or ``comparisons" against all other alternatives.
Pairwise comparisons are stochastic: to each possible pair of alternatives $(a_{i},a_{j})$ there is an associated random variable $X_{i,j} \sim P(X_{i,j})$ with finite, but unknown, expectation $\mu_{i,j}$, and making a pairwise comparison of $(a_{i},a_{j})$ is equivalent to drawing a sample from $P(X_{i,j})$. Performance of an alternative $a_{i}$ is defined to be the weighted sum of the expectation of all random variables corresponding to comparing $a_{i}$ against other alternatives in $\mathcal{P}$, i.e:

\[ S_{i} = \sum_{j\neq i} w_{i,j} \mu_{i,j} \]

In this work, we consider only the equal weighting case, where $S_{i}$ is equivalent to the simple sum of expectations or \textit{Borda Score} \citep{borda} of $a_{i}$. The top-k subset we wish to identify is defined in terms of the alternative scores as the index set $\mathcal{I}^{\star} \subset [K]$ containing the indices of the k alternatives in $\mathcal{P}$ with highest Borda score:
 \[ \mathcal{I}^{\star}  = \underset{\mathcal{I} \subset  [K]: |\mathcal{I}| = k}{argmax} \sum_{i \in \mathcal{I}} S_{i}\]

The top-k identification problem we consider is an \textit{Active Ranking} problem, whereby we sequentially collect information by selecting which pairs to sample in order to learn estimates of the pairwise means $\mu_{i,j}$ and therefore the alternative scores to inform our choice of index set. Sampling a pair is generally assumed to be expensive (perhaps in time, money or computational resources required depending on context) and so the number of possible samples we can collect is limited. Given a sampling budget, our task is to allocate samples efficiently to maximize the probability of correctly identifying the top-k subset.

As we shall discuss in Section \ref{rw}, the majority of current literature on ranking and selection using pairwise comparisons considers the case where the $X_{i,j}$ are Bernoulli R.Vs and the pairwise means $\mu_{i,j}$ correspond to the probability that alternative $a_{i}$ ``wins'' a comparison against alternative $a_{j}$. This is a useful case with many applications, for example in ranking game players when only win/loss game feedback is available, or in online advertising and search engine optimization where a user may be presented with a number of alternatives (web links), and provides binary feedback (clicks on one link). In the Bernoulli case, the pairwise random variables are generally linked by the \textit{shifted skew-symmetry} condition \citep{shah2016a}, i.e $X_{i,j} = 1 - X_{j,i}$. However, in many applications the pairwise comparison result may yield more information than a simple win/loss. For example in many pairwise games, comparisons are scored, with the score indicating the magnitude of the advantage or preference for the winning alternative. In particular, we concentrate on the common case of scored zero-sum pairwise outcomes, where the score or amount won by an alternative from a comparison is equal to the amount lost by the other (for example, 2-player Texas Hold'em poker or any other even-odds gambling game). In this scenario, the $X_{i,j}$'s are continuous, unbounded, and paired \textit{skew-symmetrically} such that $X_{i,j} = -X_{j,i}$ (zero-sum condition).

\section{Related Work}\label{rw}

The \textit{Dueling Bandits} problem was proposed in \cite{yue2009} and \cite{yue2012k} as a variation of the standard stochastic Multi-Armed Bandit (MAB) class of problems, named after an imagined multi-armed bandit casino machine. In the standard MAB problem, a decision maker sequentially acts by selecting an alternative from an available set (pulls a particular arm) and receives a noisy reward signal in return. In contrast, actions in the Dueling bandit setting consist of pairwise comparisons, where the decision maker selects a pair of alternatives (pulls two arms simultaneously) and receives some noisy feedback relating to the pair. In the standard MAB problem, reward signals are typically numerical, but in the Dueling Bandit case, pairwise feedback is generally assumed to be qualitative representing a noisy expression of preference over the two alternatives (for example ``Arm A wins against Arm B"), but quantitative feedback like score information may also be available (for example ``Arm A beats Arm B by 15 points"). As in the standard setting, the decision maker may have one of several different objectives -- for example either to maximize the total cumulative reward they can obtain over a fixed number of actions (known as the regret minimization problem), or to identify the best single arm (top-1 selection), best subset of arms (top-k selection) or to provide a complete ranking of the arms, either with the highest possible degree of confidence given a fixed sampling budget or to a pre-specified confidence level whilst minimizing the number of samples taken. In some cases it may suffice to find an alternative that is close to optimal, i.e with fitness value within some $\epsilon$ of the true best. This is referred to as the Probably Approximately Correct (PAC) setting. Each possible objective presents a different challenge, with different applications and methods. \cite{bf2018} provide a detailed review of a wide range of current literature for each problem variant for Dueling Bandit problems with qualitative feedback.
% In the regret minimization case, the decision maker must balance the trade-off between exploration and exploitation, by choosing to sample pairs that provide new information about the fitness of different alternatives, thus leading to better future sampling choices, whilst simultaneously preferentially sampling alternatives that appear better to keep total regret low. Several different approaches for identifying the best single alternative whilst minimizing regret have been proposed, for example \cite{yue2012}, \cite{yue2011}, \cite{zoghi2014}, \cite{zoghi2015} 
\subsection{Top-k Selection}
Several Dueling Bandit works focus in particular on top-k selection. \cite{bf2013} present a preference-based racing (PBR) method for active top-k selection. The objective of the PBR method is to identify the top-k subset with probability at least $1-\delta$ while minimizing the number of samples taken. To do this, it maintains an active subset of alternative pairs that are all sampled at each iteration, maintaining estimates of the win probabilities of each pair based on sampling data. When a sufficiently high degree of confidence is obtained about the pairwise sampling means of an alternative (to know with high probability whether it is in the top-k or not), the alternative is eliminated from the active set and no more samples are allocated to it, thus reducing the sampling complexity. The degree of confidence for pairwise win rates is estimated by constructing confidence intervals based on the Hoeffding Bound \citep{hoeffding1963}. In the paper, the authors present sampling strategies for 3 different ranking methods: the Copeland ranking, the random walk ranking and the sum of expectations (Borda score) ranking. While the former two are only applicable for binary comparison outcomes, the PBR sum of expectations sampling strategy would also be appropriate for our quantitative sampling case. However, the authors note that in cases where the support of the distribution of pairwise sample outcome is substantially greater than the variance, using the Empirical Bernstein bound \citep{audibert2007} to construct confidence intervals may provide better performance.

\cite{heckel2016} also suggest a racing-like sampling method for both top-k selection and total ordering of alternatives by Borda score estimation. In contrast to the PBR method, their Active Ranking (AR) method compares each alternative still in the race to a single randomly chosen opponent at each iteration, using the sample results to update alternative score estimates. They construct bounds around these estimates, again based on the Hoeffding bound, and use these to partition the alternatives into the top set and remainder. To achieve a full ranking, they use the same procedure, with $K-1$ rather than a single partition.

\cite{Mohajer2017} consider active top-k selection and top-k ranking (where the top-k subset be returned in rank order) for preference-based dueling bandits using a 2-step sampling proceedure. They first propose the SELECT algorithm for finding the top-1 alternative. SELECT uses a single-elimination tournament with repeated comparisons to counteract noise. They then generalize to the TOP algorithm for top-k ranking by dividing the entire set of alternatives into k sub-populations, and applying SELECT to find the best alternative in each. This shortlist of alternatives is then ranked and stored in order and the top alternative of the ranked shortlist is selected and removed from the list. To find the next best, only the sub-population to which the selected alternative originally belonged needs to be re-sampled, and the new top alternative from this sub-population inserted into the ranked shortlist. This process is repeated until k alternatives have been selected. SELECT/TOP assumes a total ordering over alternatives, with every pairwise preference (i.e the sign of $\mu_{i,j}$) corresponding correctly to the true alternative ranking. The single elimination tournament SELECT phase exploits this assumption by allowing alternatives to be removed from consideration based on results against only a single peer. This results in good sampling budget scaling with the number of alternatives ($\mathcal{O}(n\log n)$), but means that even a single pairwise intransitivity could make it impossible for the method to find the exact top-k, even when the number of replications per comparison in the tournament phase tends to infinity. In the paper, the authors present bounds on the samples required by the TOP method, and quantify the gain in terms of reduced sampling complexity of using active sampling over passive methods for different noise models. They compare the performance of SELECT/TOP against the AR method of \cite{heckel2016} and show superior performance on both top-1 and top-k selection.

\cite{groves2018} adapt two different Bayesian active sampling frameworks from the Simulation Optimization community for top-k selection in MAB problems with the objective of minimizing sampling complexity. The first method, Pairwise Knowledge Gradient (PKG), aims to myopically maximize the value of information \citep{frazier, branke} gained under the assumption that the chosen sample will be the final pair selected. With this assumption, potential sample pairs are rejected unless they have a high probability of the sample outcome changing the current top-k ranking. In contrast, the Pairwise Optimal Computing Budget Allocation (POCBAm) method (adapted from the non-pairwise OCBA proposed in \cite{chen1996} and later extended for subset selection in \cite{chen2008}) allocates each sample to maximize the expected increase in confidence that the current top-k ranking obtained from previous samples is correct. The authors prove that both PKG and POCBAm are asymptotically optimal when sample result distributions have infinite support, but that PKG may fail to select samples in cases where sampling outcomes are bounded. In empirical testing, they show that POCBAm outperforms both the PBR \citep{bf2013} and AR \citep{heckel2016} methods on test problems with both quantitative and qualitative sample results for several different noise models.

Our work in this paper should be seen as an extension of the POCBAm method of \cite{groves2018}, to include and exploit a parametric preference model. As the SELECT/TOP and POCABm are the best performing previously published methods, we shall compare our proposed method against both in our empirical testing.

\section{Maximum Likelihood Pairwise Optimal Computing Budget Allocation}\label{pocbam}
The general version of POCBAm proposed in \cite{groves2018} treats the results of pairwise comparisons of each distinct pair as independent. It produces a top-k ranking of alternatives by Borda score, which is equivalent to ranking by expected pairwise comparison outcome against a randomly chosen opponent. The Borda score ranking is general in that it doesn't assume stochastic transitivity conditions on pairwise comparison means. % and even if a stochastically transitive model of alternative fitness exits, the Borda score ranking is guaranteed to be identical to this ranking, so long as the empirical pairwise mean estimates obtained from sampling are sufficiently accurate. 
However, in many real-world pairwise ranking cases, (stochastic) transitivity assumptions are reasonable, 
%In such cases, particularly when there is a model available for generating these transitive relationships,
and it may be beneficial to adapt the fitness estimate to take advantage of the additional information gain from exploiting transitivity.

With this aim in mind, we propose modifying POCBAm by adding an additional assumption of a latent variable model for the pairwise outcome distributions. In particular, for the problem described in Section \ref{pd}, with continuous, skew-symmetric and unbounded pairwise comparison outcomes, we assume a Thurstonian style \citep{thurstone} model in which the fitness of each alternative $a_{i}$ is wholly determined by the value of some underlying ``quality" parameter $\gamma_{i}$. We model each pairwise comparison distribution $P(X_{i,j})$ with a Gaussian, with the mean specified by the difference $(\gamma_{i} - \gamma_{j})$ of the quality parameters of the alternatives being compared, and variance $\sigma_{i,j}^{2}$. Similar Thurstonian models have been applied to preference-based Duelling Bandits, for example in \cite{adler1994selection} or \cite{shah2016a}.

Given a set of sampling results $\{r^{(t)}_{i,j}\}^{T}_{t=1}$ of alternative pairs $(a_{i},a_{j})$, the likelihood of a set of model parameter estimates $\Gamma = (\tilde{\gamma}_{1},...,\tilde{\gamma}_{K})$ (underlying quality values) and $\Sigma = \left(\tilde{\sigma}^{2}_{i,j}\right)$ (pairwise variances) is given by:
\begin{align}
         L(\Gamma,\Sigma) &= \prod_{t=1}^{T}\frac{1}{\tilde{\sigma}^{(t)}_{i,j} \sqrt{2\pi}}e^{-\frac{1}{2}\left(\frac{-r^{(t)}_{i,j}- \tilde{\gamma}_{i}^{(t)} + \tilde{\gamma}_{j}^{(t)}}{\tilde{\sigma}^{(t)}_{i,j}}\right)^{2}} \\
        ll(\Gamma,\Sigma) &= -\frac{T}{2}log(2\pi) - \sum_{t=1}^{T}log(\tilde{\sigma}^{t}_{i,j}) - \frac{1}{2}\sum_{t=1}^{T}\left(\frac{r^{(t)}_{i,j} - \tilde{\gamma}^{(t)}_{i}+\tilde{\gamma}^{(t)}_{j}}{\tilde{\sigma}^{(t)}_{i,j}} \right)^{2}
\end{align}
Taking partial derivatives w.r.t the parameters:
\begin{align}
\begin{split}
    \fracpartial{ll(\Gamma,\Sigma)}{\gamma_{k}} &= -\sum_{t=1}^{T}\{i^{(t)}=k \}_{\mathbf{I}}\frac{ -r_{i,j}^{(t)} + \tilde{\gamma}^{(t)}_{i} - \tilde{\gamma}_{j}^{(t)}}{(\tilde{\sigma}_{i,j}^{(t)})^{2}} - \sum_{t=1}^{T}\{j^{(t)}=k \}_{\mathbf{I}}\frac{r_{i,j}^{(t)} - \tilde{\gamma}^{(t)}_{i} + \tilde{\gamma}_{j}^{(t)}}{(\tilde{\sigma}_{i,j}^{(t)})^{2}} \\
    \fracpartial{ll(\Gamma,\Sigma)}{\tilde{\sigma}_{k,l}} &= -\sum_{t=1}^{T}\frac{\{i^{(t)},j^{(t)} = k,l\}_{\mathbf{I}}}{\tilde{\sigma}_{i,j}^{(t)}} + \sum_{t=1}^{T}\{i^{(t)},j^{(t)} = k,l\}_{\mathbf{I}}\frac{\left(r^{(t)}_{i,j}- \tilde{\gamma}^{(t)}_{i}+\tilde{\gamma}^{(t)}_{j}\right)^2}{(\tilde{\sigma}^{(t)}_{i,j})^{3}}
\label{ll_eq}
\end{split}
\end{align}
Where $\{i^{(t)} = k\}_{\mathbf{I}}$ is the 0/1 indicator function returning 1 when the first alternative in the pair sampled in the $t^{th}$ sample is $a_{k}$, $\{j^{(t)} = k\}_{\mathbf{I}}$ the analogous function for the second alternative, and $\{i^{(t)},j^{(t)} = k,l\}_{\mathbf{I}}$ the indicator function that returns 1 when the pair of alternatives in the $t^{th}$ sample were $a_{k}$ and $a_{l}$ in either order. Calculating these partial derivatives, we can then maximize the (log) likelihood of the model parameters given our collected sampling data using a quasi-newton method.

We initialize the model parameter vector $\Gamma$ with the alternative Borda Score estimates, centered (arbitrarily) about  $\tilde{\gamma}_{0}$ and divided by population size, i.e:
\[\tilde{\gamma}_{i} = \frac{1}{K}\left(\sum_{j \neq i} \tilde{\mu}_{i,j} - \sum_{j \neq 0} \tilde{\mu}_{0,j}\right)\]
and the pairwise variance parameters $\Sigma$ with the sample standard deviations:
\[\tilde{\sigma}_{p,q}^{2} = \tilde{\sigma}_{p,q}^{2} = \frac{1}{n_{p,q}-1}\sum_{t=1}^{T} (r^{(t)} - \tilde{\mu}_{p,q} )^{2} \{i^{(t)},j^{(t)} = p,q \}_{\mathbf{I}}\]
These  values represent a reasonable initial guess at the model parameters without taking account of the inter-dependence of pairwise sampling results. We then perform gradient based optimization of the parameter log likelihood (Equation \ref{ll_eq}) to converge to a local maximum, providing a better set of parameter estimates from the whole data, whilst exploiting the structure of the proposed underlying preference model. To ensure that the starting parameter estimates are reasonable when first beginning the sampling process, we allocate a portion of the sampling budget uniformly across all alternative pairs, taking $n_{0}$ samples of each.

After this initial ``warm-up" sampling phase, we use this sampling information collected to inform our active sample selection procedure. To sample efficiently, we sequentially allocate additional samples to myopically maximize the expected increase in our confidence that the data we have collected allows us to correctly identify the  top-k subset of alternatives. To quantify our level of confidence in our current top-k subset, we estimate the probability that the alternatives with the highest $k$ Borda score estimates based on the current sampling information are indeed the true best k. As in \cite{chen1996} and \cite{chen2008}, we refer to this as the Probability of Correct Selection (PCS). For a given index set $\mathcal{I}$ of the current top alternatives, this is defined as:

\[PCS= \mathbb{P}\{ S_{i} > S_{j} ,\quad  \forall i \in \mathcal{I}, j \notin \mathcal{I} \}\]

Under the assumption of normality of pairwise sample mean estimates, we can construct approximated posterior distribution estimates for the alternative Borda scores using the fitted model parameter estimates $\Gamma^{*} = [\gamma^{*}_{1},...,\gamma^{*}_{K}], \Sigma^{*} = \left( (\sigma^{*}_{i,j})^{2} \right)$  obtained from the optimization step above:

\[\hat{S}_{i} \sim \mathcal{N}\left[\bm{\hat{\mu}_{i}}, \bm{\hat{\sigma}_{i}}^{2}\right] = \mathcal{N}\left[ \sum_{j \neq i} (\gamma^{*}_{i} - \gamma^{*}_{j}), \sum_{j \neq i} \frac{(\sigma^{*}_{i,j})^{2}}{n_{i,j}} \right] \]

As in \cite{groves2018}, we simplify the calculation of PCS with two approximations. Firstly, we use a threshold value to separate the score distributions of the current top-k alternatives from the rest of the population. For a constant $c$,

\[PCS \geq \mathbb{P} {\left[ \left( \bigcap_{p \in \mathcal{I}} \{ \hat{S}_{p}  > c \} \right) \bigcap \left( \bigcap_{q \notin \mathcal{I}} \{ \hat{S}_{q} < c \} \right)\right]\equiv APCS} \label{pcs_ineq}\]

Ideally we want to select $c$ carefully to make this lower bound as accurate as possible. \cite{chen2008} show that APCS is asymptotically maximized for

\[c = \frac{\bm{\hat{\sigma}_{k +1}}\bm{\hat{\mu}_{k}} + \bm{\hat{\sigma}_{k}}\bm{\hat{\mu}_{k+1}}}{\bm{\hat{\sigma}_{k}}+\bm{\hat{\sigma}_{k +1}}} \label{c} \]

Where $\bm{\hat{\mu}_{k}}$ and $\bm{\hat{\mu}_{k + 1}}$ are the means and $\bm{\hat{\sigma}_{k}}$ and $\bm{\hat{\sigma}_{k+1}}$ are the standard deviations of the current k\textsuperscript{th} and (k+1)\textsuperscript{th} best alternatives respectively. Secondly, evaluating the intersections in Equation \ref{pcs_ineq} directly is difficult as the alternative score distributions are not independent. Instead we use the following approximation:

\[ APCS \approx \prod_{p \in \mathcal{I}} \mathbb{P} \{ \hat{S}_{p}  > c \} \prod_{q \notin \mathcal{I}} \mathbb{P} \{ \hat{S}_{q} < c\}\]

The justification for this approximation is discussed in detail in Section 4.1 of \cite{groves2018} (Inequalities 4.1.3 and 4.1.4). Using these approximations reduces the complexity of calculating PCS from $\mathcal{O}(K^{2})$ to $\mathcal{O}(K)$, which is useful as our active sample selection procedure calculates predicted post-sample AEPCS for each of the $(K^{2}-K)/2$ possible alternative pairs.

To inform our sample selection, we predict the approximate alternative score distributions if we were to allocate an additional sample to a particular pair. The expected effect of this sample would be a reduction in the uncertainty of the estimate of the pairwise mean $\mu_{i,j}$, thereby increasing our APCS estimate. It is important to note that, unlike the standard version of POCBAm described in \cite{groves2018}, the score distribution parameters obtained from likelihood maximization are no longer unbiased estimates, and increasing the confidence in the sample estimate of a single pairwise mean would also affect the likelihood of the entire set of fitted model parameters. To avoid repeating the costly optimization step for every possible alternative pair before selecting each sample, we make a further approximation by restricting the effect to only the score distributions of the two alternatives in the sampling pair. The potential error introduced by this approximation should be mitigated in part by the fact that we only require relative APCS values and by the performance benefit from exploiting preference transitivity. In Section \ref{tests}, we demonstrate that despite this simplification, ML-POCBAm is still able to achieve leading performance on a range of empirical tests.

Under our earlier assumption of normality, for a sample allocated to the pair $(a_{p},a_{q})$, our expected score distributions are:

\[\hat{S}^{p,q}_{i} \sim \mathcal{N}\left[ \sum_{j \neq i} (\gamma^{*}_{i} - \gamma^{*}_{j}), \sum_{j \neq i} \frac{(\sigma^{*}_{i,j})^{2}}{n_{i,j} + \{i,j = p,q \}_{\mathbf{I}} } \right]\]

\begin{figure}
    \centering
    \includegraphics[width=\textwidth]{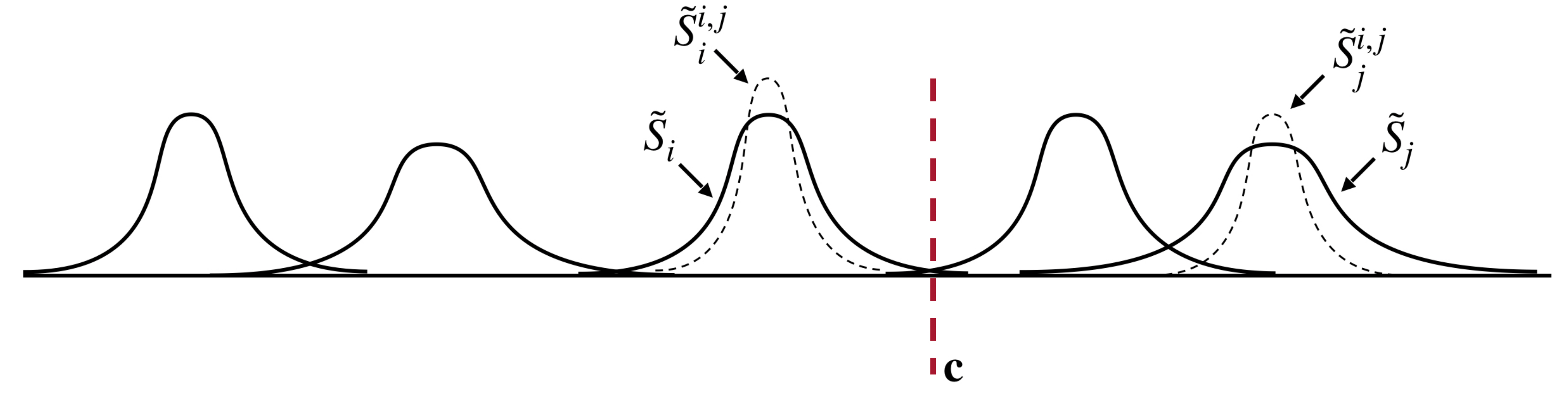}
    \caption{Illustration of the expected effect on the approximated posterior score distributions due to allocating a sample to the pair $(a_{i},a_{j})$. The expected post-sample distributions $\hat{S}^{i,j}_{i}$ and $\hat{S}^{i,j}_{j}$ are narrower, increasing the probability mass lying on the ``correct" side of the threshold $c$ and thereby increasing AEPCS.}
    \label{fig:ocba_illustration}
\end{figure}

Figure \ref{fig:ocba_illustration} gives an illustration of the expected effect of a sample on the score distributions relative to the threshold value $c$. To choose which sample to take, our active sampling method calculates an expected value for $APCS$ for each potentially sampled pair $(a_{i},a_{j})$, denoted by $AEPCS^{i,j}$, and samples wherever this quantity is maximized:
\begin{align}
\begin{split}
AEPCS^{i,j} &= \prod_{p \in \mathcal{I}} \mathbb{P}\{\hat{S}^{i,j}_{p} > c \}\prod_{q \notin \mathcal{I}} \mathbb{P}\{\hat{S}^{i,j}_{q} < c \} \\
&={\displaystyle \prod_{p \in \mathcal{I}}} \left(1-\Phi \left(\frac{c-\bm{\hat{\mu}^{i,j}_{p}}}{\bm{\hat{\sigma}^{i,j}_{p}}}\right)\right){\displaystyle \prod_{q \notin \mathcal{I}}} \left(\Phi \left(\frac{c-\bm{\hat{\mu}^{i,j}_{q}}}{\bm{\hat{\sigma}^{i,j}_{q}}}\right)\right)
\end{split}
\end{align}

Where $\Phi$ is the cumulative distribution function for the standard normal distribution and using $c$ as defined in Equation \ref{c}. A step-by-step description of the method is shown in Table \ref{alg}.

\begin{table}
\caption{}
\vspace{5pt}
\begin{tabular}{llll}
\hline
\multicolumn{4}{l}{\textbf{POCBAm Procedure}} \\
\hline
\multicolumn{2}{l}{INPUT:} & \multicolumn{2}{l}{Set of $K$ alternatives $\{a_{1},..,a_{K}$\}, }\\
&& \multicolumn{2}{l}{Required selection size $k$,} \\
&& \multicolumn{2}{l}{Sampling budget $N$.}\\
\multicolumn{2}{l}{INITIALIZE:} & \multicolumn{2}{l}{Perform $n_{0}$ samples of each pair of alternatives;} \\
&& \multicolumn{2}{l}{$n_{p,q} = n_{0}$ for all $p,q$,} \\
&&\multicolumn{2}{l}{Sample means $\tilde{\mu}_{p,q} = \frac{1}{n_{p,q}} \sum X_{p,q}$, and:}\\
&& \multicolumn{2}{l}{Standard dev. $\tilde{\sigma}_{p,q} = \sqrt{\frac{1}{n_{p,q}-1}\sum (X_{p,q} - \tilde{\mu}_{p,q})^{2}}$,}\\
%&& \multicolumn{2}{l}{For all $p = 1,...,K$: alternative score estimates $\hat{S}_{p} = \sum_{q,q\neq p}\tilde{\mu}_{p,q}$,}\\ 
&&\multicolumn{2}{l}{Index set $\mathcal{I}$ of best $k$ alternatives.} \\\\
\multicolumn{4}{l}{WHILE $\sum n_{i,j} < N$ DO:} \\
&\multicolumn{3}{l}{FOR ALL PAIRS $(a_{i},a_{j})$:} \\
&&UPDATE:&\\
&&Initial model parameter estimates:&\\
&&\hspace{15pt} $\tilde{\gamma}_{p} = \frac{1}{K}\left(\sum_{q \neq p} \tilde{\mu}_{p,q} - \sum_{q \neq 0} \tilde{\mu}_{0,q}\right)$,\\
&&\hspace{15pt} $\tilde{\sigma}_{p,q}^{2} = \tilde{\sigma}_{p,q}^{2} = \frac{1}{n_{p,q}-1}\sum_{t=1}^{T} (r^{(t)} - \tilde{\mu}_{p,q} )^{2} \{i^{(t)},j^{(t)} = p,q \}_{\mathbf{I}}$,&\\
&& Fit parameter estimates: &\\
&&\hspace{15pt} $(\Gamma^{*}, \Sigma^{*}) = (\gamma^{*}_{0},...\gamma^{*}_{K}), \left( \sigma^{*}_{i,j} \right)= \text{argmax}[ll(\Gamma,\Sigma)]$&\\
&&For all $\bm{p} = 1,..,K$:&\\
&&\hspace{15pt}$\hat{S}_{i} \sim \mathcal{N}\left[\bm{\hat{\mu}_{i}}, \bm{\hat{\sigma}_{i}}^{2}\right] = \mathcal{N}\left[ \sum_{j \neq i} (\gamma^{*}_{i} - \gamma^{*}_{j}), \sum_{j \neq i} \frac{(\sigma^{*}_{i,j})^{2}}{n_{i,j}} \right]$&\\
&&\hspace{15pt} Alternative score means $\bm{\hat{\mu}^{i,j}_{p}} := \bm{\hat{\mu}_{p}}$,&\\
&&\hspace{15pt} Alternative score variances $(\bm{\hat{\sigma}^{i,j}_{p}})^{2} :=\sum_{q, q\neq p} \frac{(\sigma^{*}_{p,q})^{2}}{n_{p,q} + \mathbb{I}\{p,q = i,j\}}$,&\\
&&Boundary value $c = \frac{\bm{\hat{\sigma}_{k +1}}\bm{\hat{\mu}_{k}} + \bm{\hat{\sigma}_{k}}\bm{\hat{\mu}_{k+1}}}{\bm{\hat{\sigma}_{k}}+\bm{\hat{\sigma}_{k +1}}}$,&\\
&&$AEPCS^{i,j}= {\displaystyle \prod_{p \in \mathcal{I}}} \left(1-\Phi \left(\frac{c-\bm{\hat{\mu}^{i,j}_{p}}}{\bm{\hat{\sigma}^{i,j}_{p}}}\right)\right){\displaystyle \prod_{q \notin \mathcal{I}}} \left(\Phi \left(\frac{c-\bm{\hat{\mu}^{i,j}_{q}}}{\bm{\hat{\sigma}^{i,j}_{q}}}\right)\right)$.&\\
&\multicolumn{3}{l}{END FOR}\\
&SAMPLE:&\multicolumn{2}{l}{Select pair $(a_{i},a_{j})$ that maximizes $AEPCS^{i,j}$,}\\
&&\multicolumn{2}{l}{Perform sample of $(a_{i},a_{j})$,}\\
&&\multicolumn{2}{l}{$n_{i,j} \leftarrow n_{i,j}+1$,}\\
&&UPDATE: $\tilde{\mu}_{i,j}, \tilde{\sigma}_{i,j}, \hat{S}_{i}, \hat{S}_{j}$,&\\
&UPDATE: & \multicolumn{2}{l}{$\mathcal{I}$.}\\
\multicolumn{3}{l}{END WHILE}\\\\
\multicolumn{2}{l}{RETURN} & \multicolumn{2}{l}{$\mathcal{I}$}\\
\hline
\label{alg}
\end{tabular}
\end{table}

\section{Empirical Testing}\label{tests}
In this section we evaluate the performance of the \textit{ML-POCBAm} method on several test scenarios generated with by varying the underlying pairwise preference model and on a realistic poker player top-k selection task for identifying the best performing subset of No Limit Texas Hold'em playing agents. We compare the method performance against the \textit{SELECT/TOP} method proposed in \cite{Mohajer2017}, the standard \textit{POCBAm} method as described in \cite{groves2018} and against uniform sample allocation. We use correct selection success rate as our performance metric for each task, defined as the proportion of correctly identified top-k subsets over a large number of independent replications of the experiment, with different alternative populations and pairwise variances:
\[Success Rate = \frac{\{\mathcal{I} = \mathcal{I}^{*} \}_{\mathbf{I}}}{N_{replications}}  \]
For the following experiments, unless otherwise stated, $N_{replications} = 10,000$ and the initial random seeds used to generate the underlying alternative parameters in each replication are common to each method (i.e the true alternative Borda scores for the $n^{th}$ replication of the ML-OCBAm method are the same as for the $n^{th}$ replication of the SELECT/TOP method).

\subsection{Value-based Thurstone model}
\begin{figure}
\begin{subfigure}{.5\linewidth}
\centering
\includegraphics[width=1.1\linewidth]{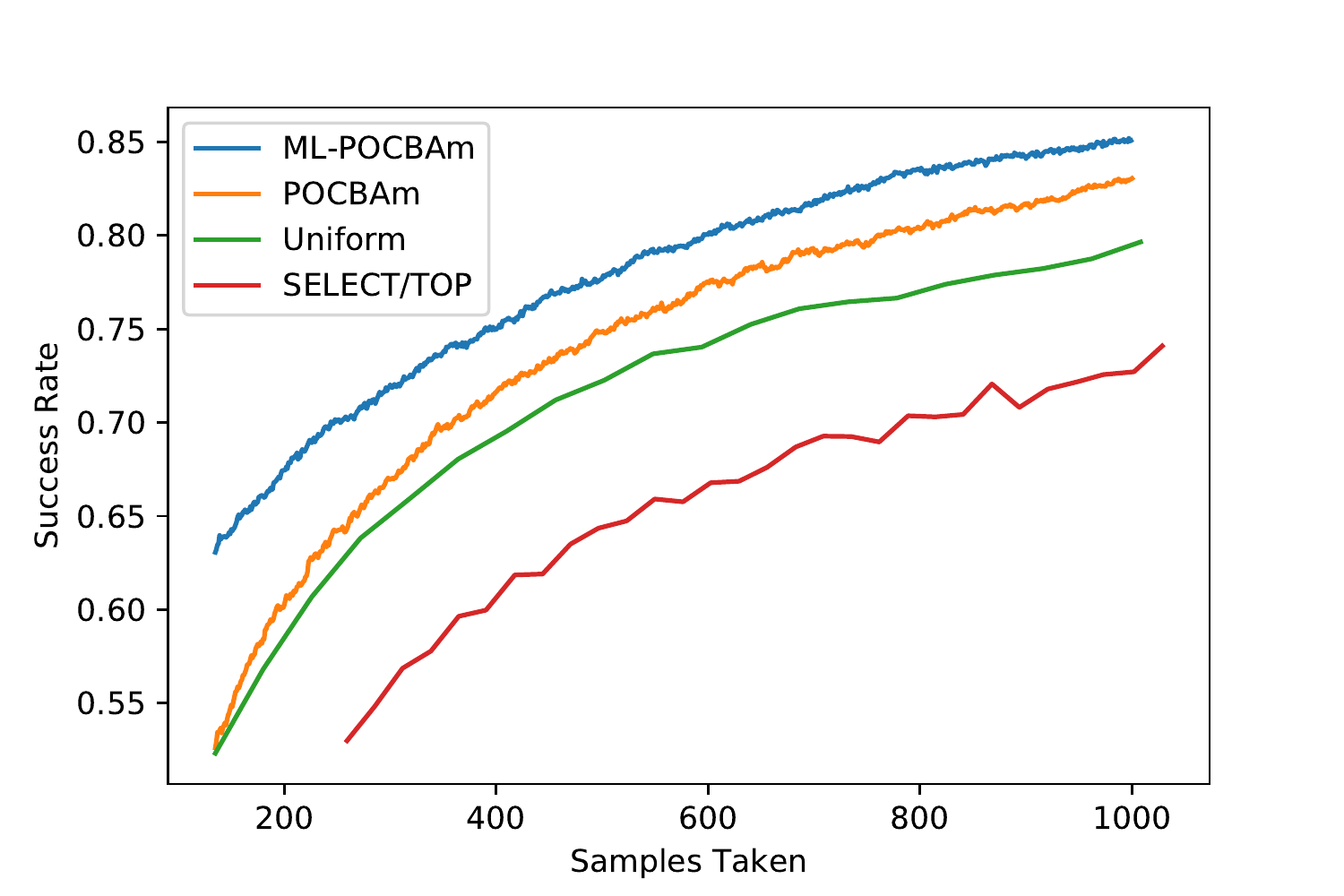}
\caption{4 of 10}
\label{f1:4of10}
\end{subfigure}%
\begin{subfigure}{.5\linewidth}
\centering
\includegraphics[width=1.1\linewidth]{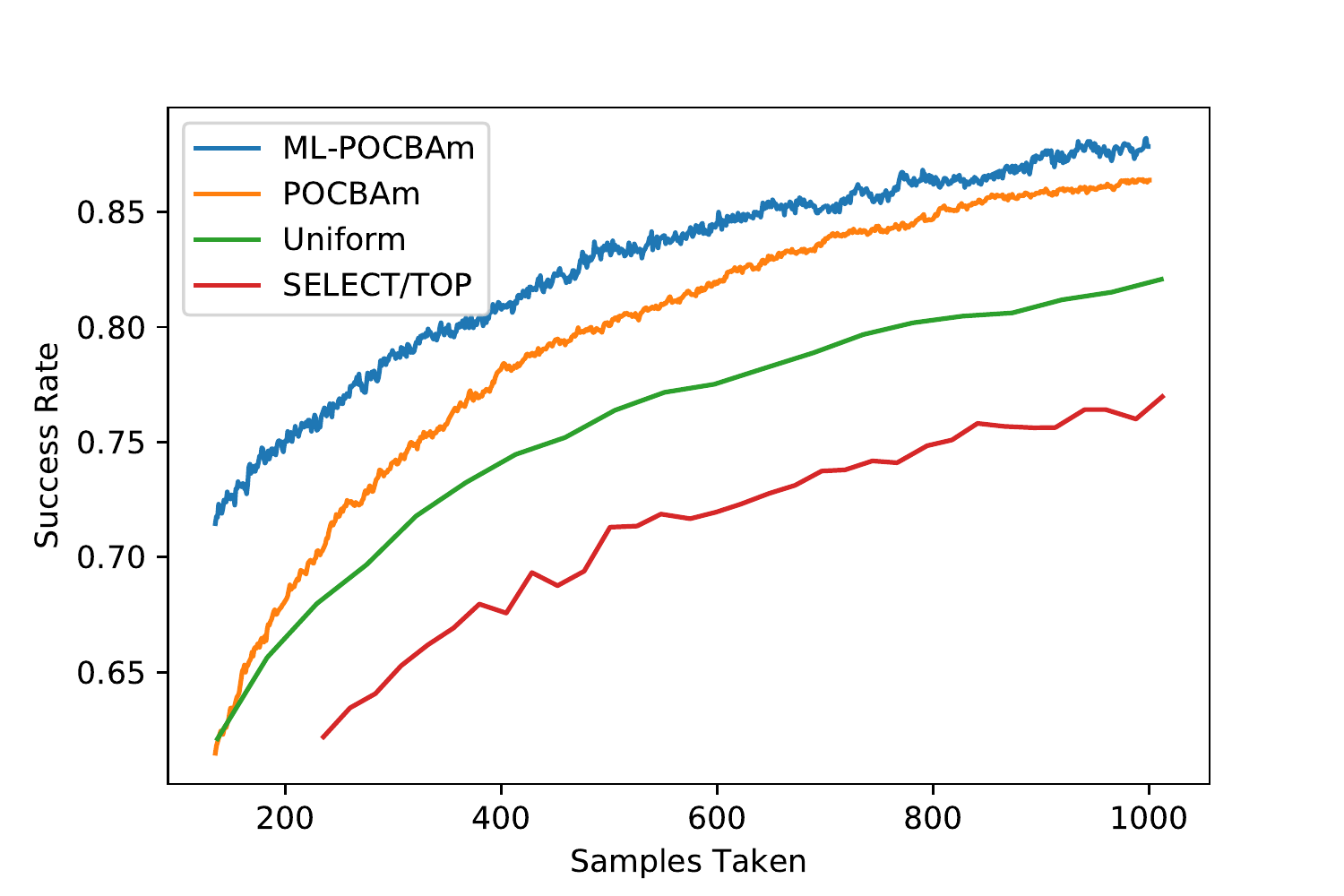}
\caption{1 of 10}
\label{f1:1of10}
\end{subfigure}
\centering
\begin{subfigure}{.5\linewidth}
\centering
\includegraphics[width=1.1\linewidth]{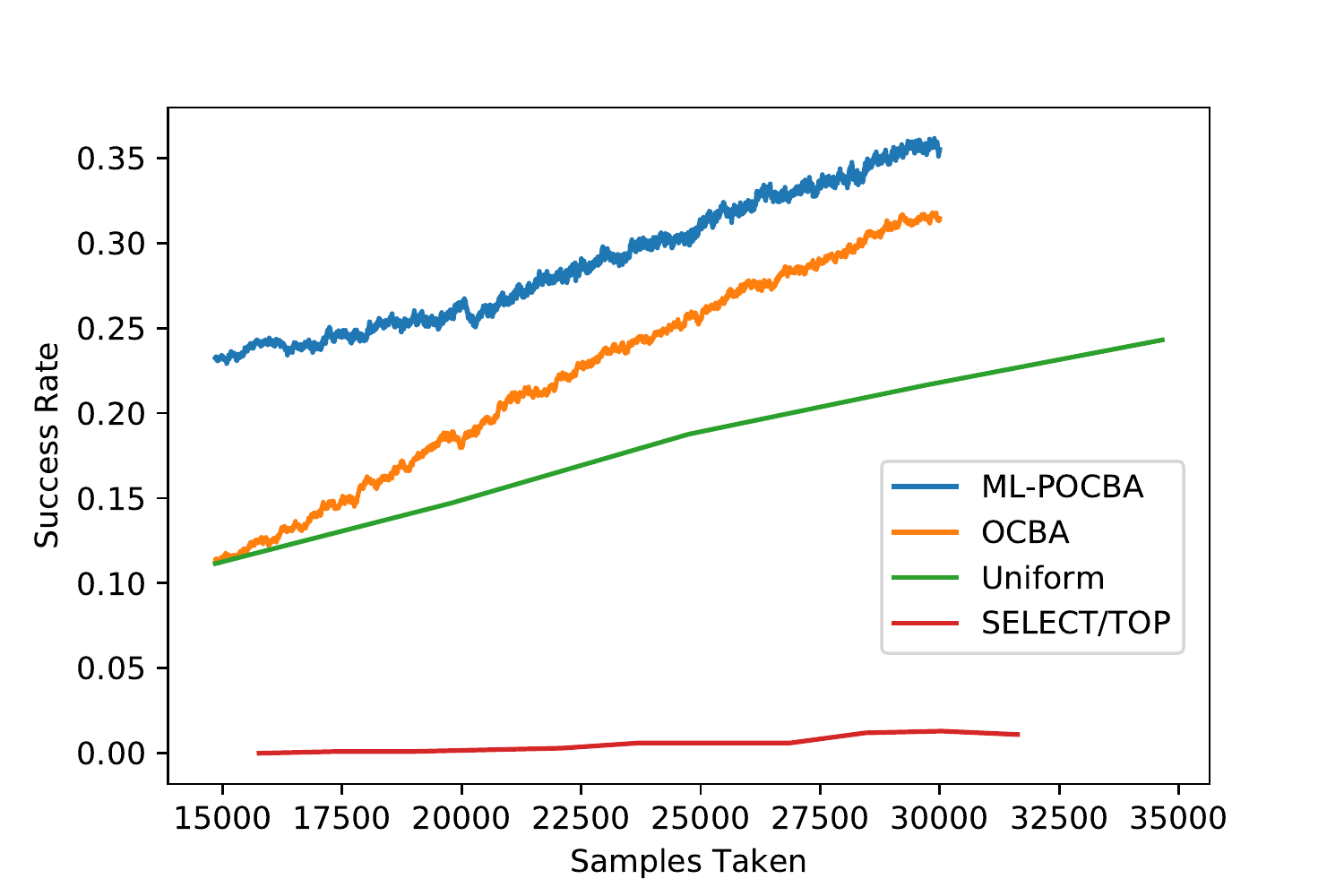}
\caption{40 of 100}
\label{f1:40of100}
\end{subfigure}
\caption{Performance of ML-POCBAm, POCBAm, SELECT/TOP and uniform sample allocation on top 4 of 10, top 1 of 10 and top 40 of 100 selection with a Thurstonian latent preference model.}
\label{f1:thurstone}
\end{figure}

The first test scenario we consider is where the pairwise preference distributions are generated according to the value-based Thurstonian model described in Section \ref{pocbam}, i.e. the underlying model assumed by ML-POCBAm is correct. Note that in this case, the total ordering assumption of the SELECT/TOP method also holds. In each replication of the experiment, the true parameter vector $\Gamma$ for our population of alternatives is generated with each entry $\gamma_{i}$ chosen uniformly at random from $[0,1]$. The pairwise variance parameters $\sigma_{i,j}^{2}$ are also selected independently and uniformly from $[0,1]$ for each pair of alternatives. These parameter values determine the true top-k ranking, and the pairwise sampling distributions. At each step $t$ of the experiment, each of the sampling methods may select a pair of alternatives $(a_{i},a_{j})$ and receive a sample result drawn from the distribution $\mathcal{N}(\gamma_{i} - \gamma_{j}, \sigma_{i,j}^{2})$. The ML-POCBAm and POCBAm methods were allowed a maximum of 1000 samples when selecting from 10 alternatives, and 30,000 when selecting from 100, with the first 3 samples for each pair allocated uniformly by each method to make the initial pairwise mean estimates. For the SELECT/TOP method, the number of samples taken by the method is variable, and is dependent on a pre-chosen parameter $\nu$ that specifies the number of repetitions of each comparison to make during the knockout tournaments and heap construction phases. As such, we vary $\nu$ to produce a range of different average sampling budgets, without limiting the maximum sampling budget used by this method. Figure \ref{f1:thurstone} shows the method performance on both top 4 of 10, top 1 of 10 selection and larger scale top 40 of 100 selection. In all three cases, we see that ML-POCBAm achieves highest success rate throughout the 1000 samples, with regular POCBAm performing second best. The benefit of exploiting the transitivity of the model is clear, as fitting the most likely model parameters given only the initial warm-up sampling data significantly improves success rate. Interestingly, SELECT/TOP performs worse than uniform sampling in each of the cases, which is surprising given its strong performance in \cite{Mohajer2017} and the required assumption of a total ordering holds. We discuss the cause for this poor performance in the following subsection.

\subsubsection{Analysing SELECT/TOP}
As \cite{Mohajer2017} note, the probability of correct selection of the SELECT tournament stage is highly dependent on the difference in quality of the top alternative and the remainder. If any of the best alternatives are knocked out of a sub tournament it is much more difficult SELECT/TOP to return them to consideration than for ML-POCBAm to produce a correct ranking given a poor estimate of a single pairwise mean. It is therefore critical to the performance of SELECT/TOP that the best alternatives are highly unlikely to ever lose in a comparison to alternatives not in the correct top set. 
Taking the above top 1 of 10 experiment as an example, we generated the alternative score parameters ($\gamma_{i}$'s) uniformly from $[0,1]$. This means that the expected comparison mean between the top and 2\textsuperscript{nd} best alternative will be the difference between the 9\textsuperscript{th} and 10\textsuperscript{th} order statistic of 10 uniform samples (\cite{gentle2009computational}, page 63): 1/11 = 0.0909. Using the expected value of the standard deviation parameter for this pair (0.5), the probability of the top alternative winning over $n$ comparisons against the second best is given by:
\[1 -  \Phi \left( \frac{n/11}{\sqrt{0.25n}} \right)\] 
Which is only about 0.572 when $n=1$ and 0.717 when $n=10$. The probability of the top alternative losing in one of the several tournament rounds is therefore significant.

\begin{figure}
    \centering
    \includegraphics{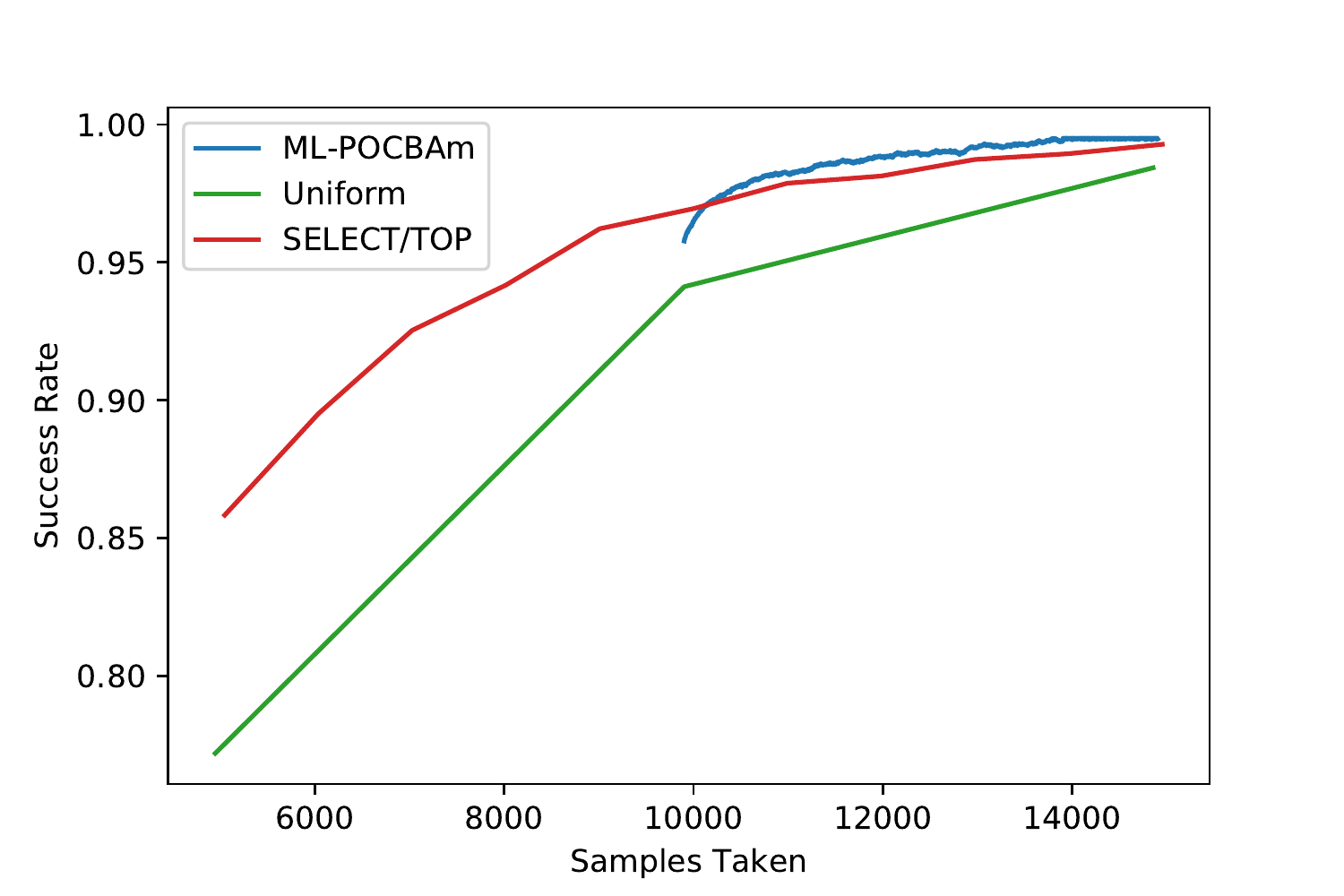}
    \caption{Performance of ML-POCBAm, Uniform sampling and SELECT/TOP on top 1 of 100 selection with a Thurstonian latent preference model and a large separation between the quality value of the top alternative and the rest of the population. Here $\gamma_{0} = 1.2$, with $\gamma_{i}$ selected uniformly at random from $[0,1]$ for all other alternatives.}
    \label{fig:1of100}
\end{figure}

However, one notable advantage of SELECT/TOP over both ML-POCBAm and POCBAm is its lower sampling complexity. To generate initial estimates for alternative score distributions, both OCBA-based methods use a small number of warm-up samples of each pair, which scales $O(n^{2})$ with the number of alternatives. In contrast, SELECT/TOP has complexity $O(n\log(n))$. As the number of alternatives grows large, the cost of the warm-up phase of ML-POCBAm will become dominant and can restrict performance.
Therefore, to ensure SELECT/TOP has a fair chance to display its merits, we compare it to ML-POCBAm and Uniform sampling using the larger population size (100) used above, selecting the top-1 alternative, where the gap in underlying quality value between the best alternative and the rest is relatively large. To do this, we set $\gamma_{0} = 1.2$ and sample $\gamma_{1:99}$ and pairwise variances $\sigma^{2}_{0:99,0:99}$ from $U[0,1]$. Results are shown in Figure \ref{fig:1of100}. We observe that the large gap between the best alternative and the remainder improves the success rates for all three methods, as the selection problem is now significantly easier. SELECT/TOP in particular performs much better here, reaching the same success rate as ML-OCBAm achieves immediately after its warm-up phase with approximately 12\% fewer samples. However ML-POCABm is still able to reach perfect accuracy earlier. After the initial warm-up phase ML-POCBAm is able to refit its model, recalculate AEPCS and thus allocate every sample to the myopically optimal pair. This leads to substantially more efficient sample acquisition than using fixed rules for the number of sample repetitions as in SELECT/TOP and uniform allocation.

\section{Noise Perturbed Thurstone model}\label{nt}
\begin{figure}
     \centering
     \includegraphics{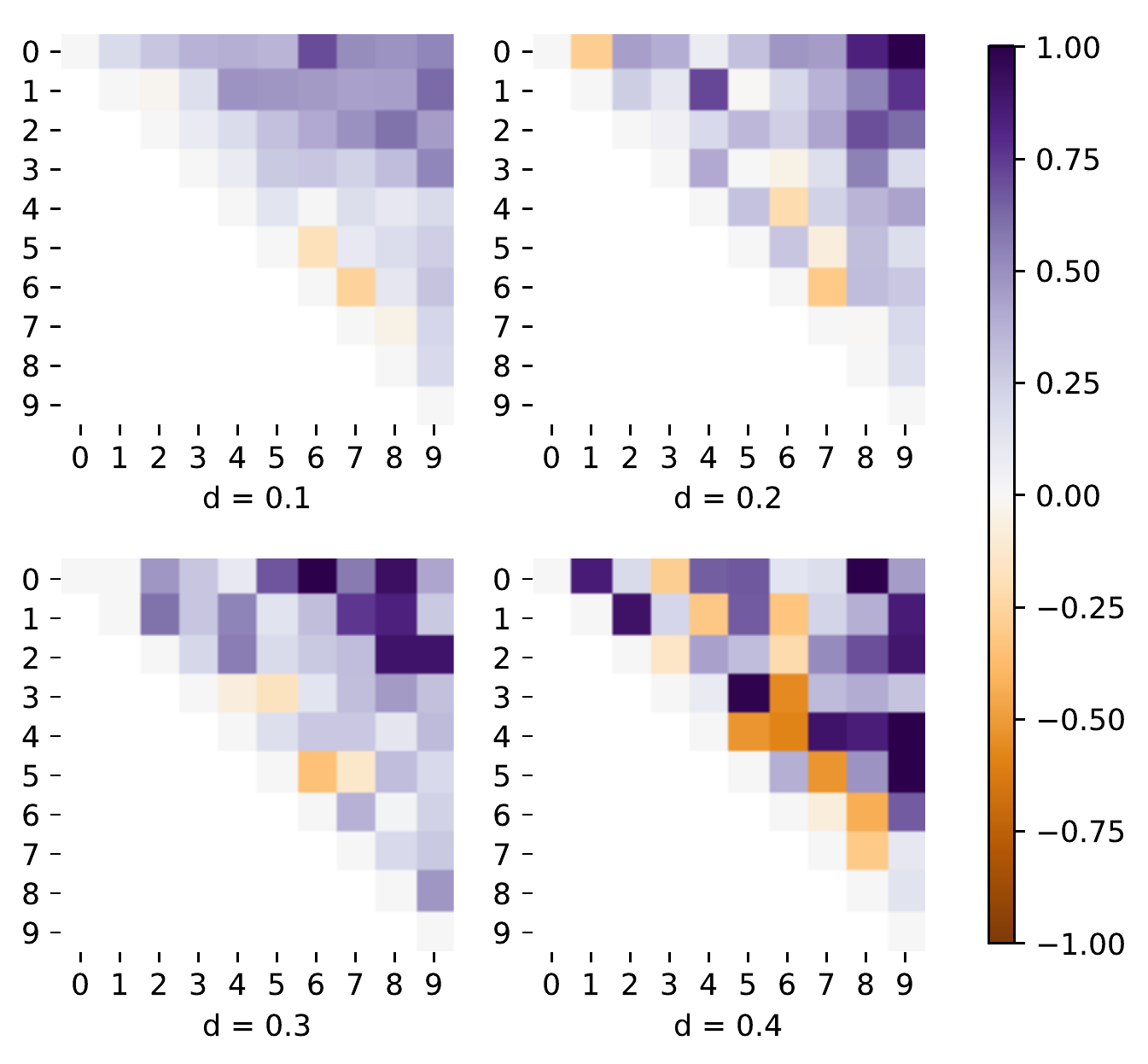}
     \caption{Sample pairwise preference matrices for populations of 10 alternatives generated according to the noise perturbed Thurstonian model described in \ref{nt} with different values of $d$. Alternatives are indexed by Borda score. We can clearly see this increasing degree of pairwise intransitivity as $d$ increases. For an example of an intransitive triplet, consider $(a_{0}, a_{1}, a_{2})$ for $d=0.4$, we have $\mu_{0,1} > 0, \mu_{1,3} > 0, \mu_{0,3} < 0$.}
     \label{fig:nt_examples}
 \end{figure}

\begin{figure}
\begin{subfigure}{.5\linewidth}
\centering
\includegraphics[width=1.1\linewidth]{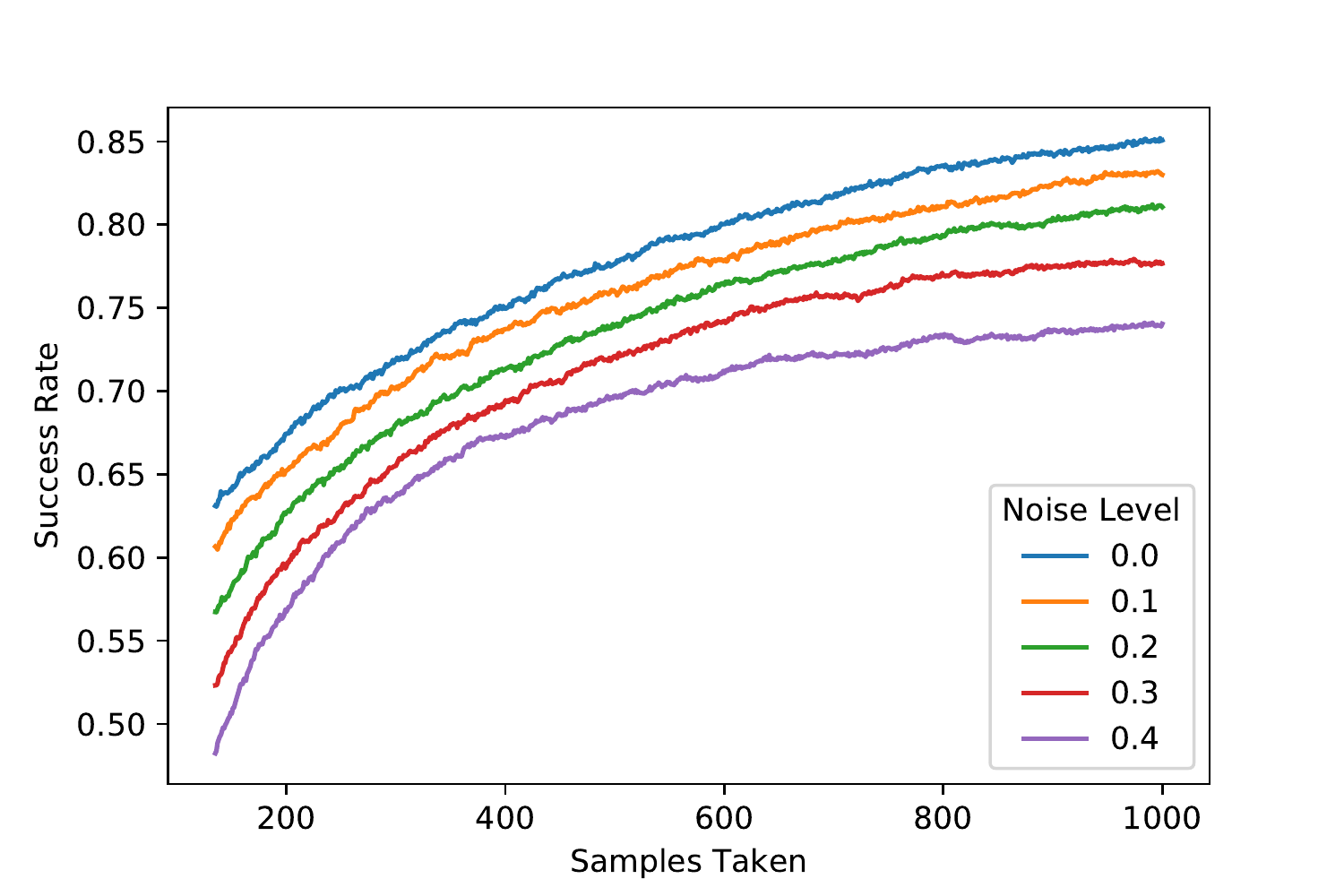}
\caption{ML-POCBAm}
\label{f2:mle}
\end{subfigure}%
\begin{subfigure}{.5\linewidth}
\centering
\includegraphics[width=1.1\linewidth]{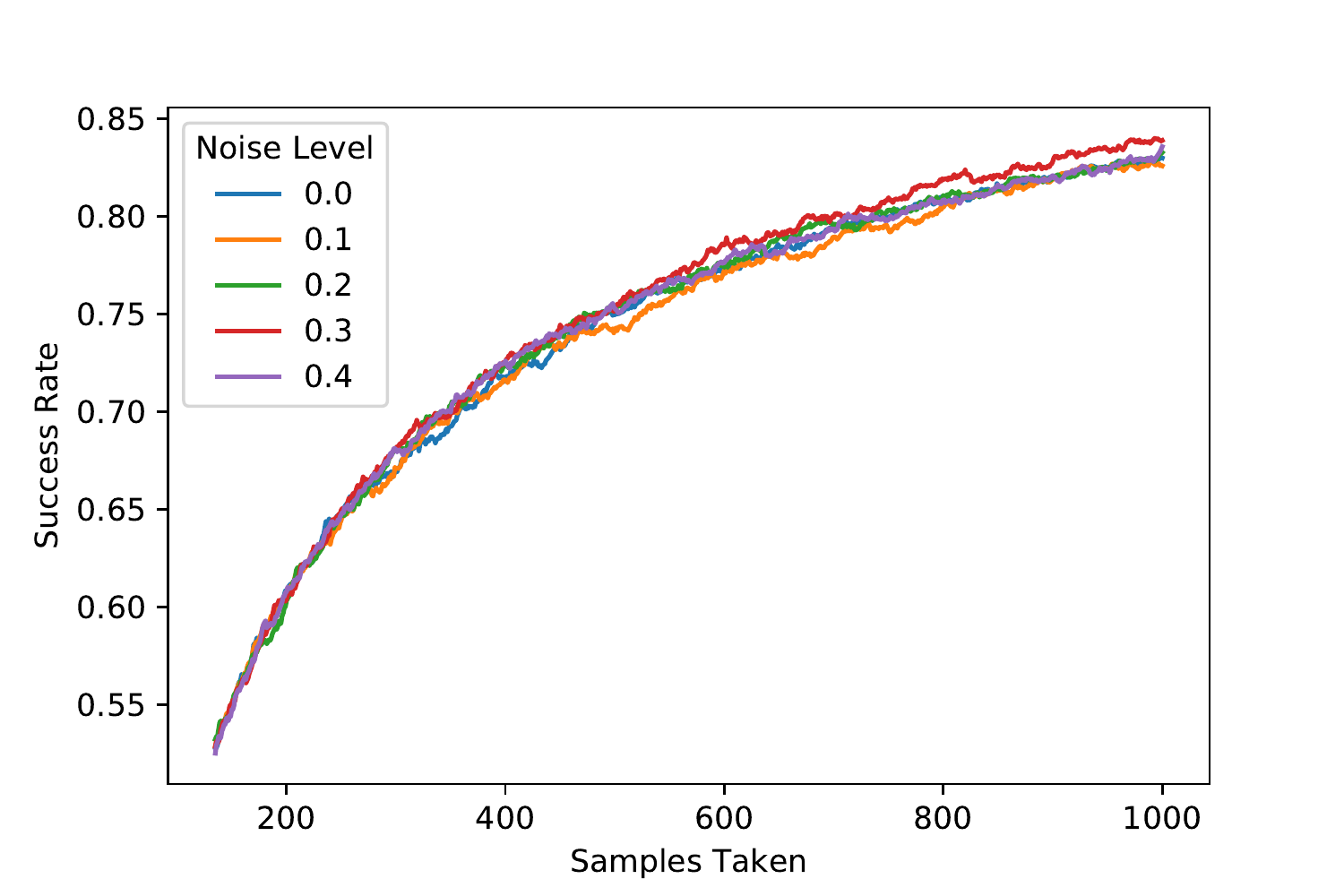}
\caption{POCBAm}
\label{f2:std}
\end{subfigure}\\[1ex]
\begin{subfigure}{\linewidth}
\centering
\includegraphics[width=0.55\linewidth]{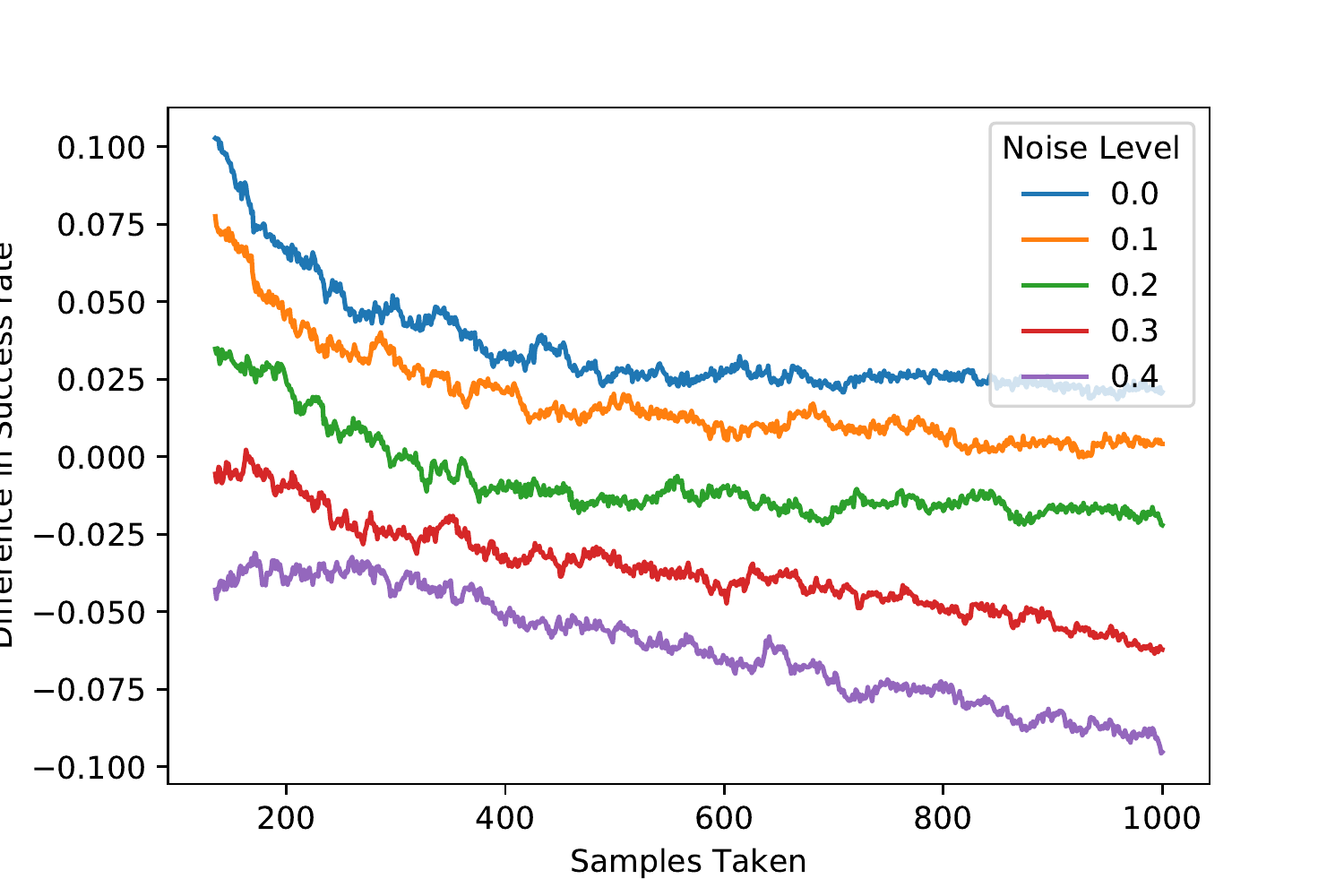}
\caption{Performance difference}
\label{f2:diff}
\end{subfigure}
\caption{Performance of ML-POCBAm and POCBAm on top 4 of 10 selection problems with noise disturbed Thurstone preference with various different degrees of noise.}
\label{fig1:test}
\end{figure}

In the experiments above, the ground-truth mechanism for generating pairwise sample outcomes corresponds exactly with the model used by ML-POCBAm. Under these conditions ML-POCBAm performs very well, but having sole access to the true model gives the method a potentially unrealistic advantage over its competitors. In a real-world scenario, the underlying model is generally unknown, and the rigid assumptions of the Thurstonian model are unlikely to hold exactly, even if they are broadly correct. In this section, we investigate the effect on algorithm performance of random perturbations to the pairwise means of the underlying Thurstonian model, and propose an adaptation to the method to make it more resilient to model inaccuracy. Here the model parameters are generated in the same way as before, but the means of the sampling distributions are each perturbed by an additive noise component, i.e, for each pair $(a_{i}, a_{j})$ samples are drawn from the distribution $\mathcal{N}(\gamma_{i} - \gamma_{j} + \epsilon_{i,j}, \sigma_{i,j}^2)$. The $\epsilon_{i,j}$'s are chosen independently at random from the distribution $\mathcal{N}(0,d^{2})$. An example of the effect of these perturbations on the pairwise transitivity of the alternatives for $d = \{0.1,0.2,0.3,0.4\}$ is shown in Figure \ref{fig:nt_examples}. As $d$ increases the number of alternative pairs that violate the total ordering assumption (any negative mean in the upper triangle of the ordered pairwise mean matrix) or the weak stochastic transitivity condition as defined in \cite{falahatgar2018limits} (alternative triplets $(a_{i}, a_{j}, a_{k})$ where $\mu_{i,j} > 0, \mu_{j,k} > 0$, but $\mu_{i,k} < 0$) increases considerably. Figure \ref{fig1:test} shows the performance of ML-POCBAm and the best performing competing method from the previous section (POCBAm) on similarly generated cases. We see that the performance of ML-POCBAm is significantly reduced as $d$ increases, and the parametric model used by ML-OCBAm becomes a poorer representation of the population. In contrast, POCBAm is resilient to the perturbations, as each pairwise mean is treated independently. Figure \ref{f2:diff} highlights the difference in performance between the two methods. For low $d$, fitting the Thurstonian model to the data is still beneficial to selection accuracy, although as sampling budget increases, this advantage lessens, as POCBAm will asymptotically converge to the correct subset. For high $d$ values, POCBAm performs better than ML-OCBAm, with the performance gap widening as more samples are taken. This suggests that when the model used by ML-POCBAm is sufficiently inaccurate, fitting this model to the sampling data harms not only prediction accuracy, but also sample acquisition.

Clearly it would be beneficial to use a sampling method that could exploit the benefit of fitting the parametric model when appropriate, but can retain the resilience to pairwise intransitivity of the POCBAm method. However, without knowing the true top-k subset, it is difficult to know what the effect of modelling inaccuracy is on correct selection. \cite{samothrakis2013} discusses different methods for estimating the degree of intransitivity when ranking pairwise systems. They suggest comparing the symmetrized Kullback Leibler (KL) divergence between the observed (sample) distributions and the predicted pairwise distributions according to the model. Using this idea, we define the empirical Intransitivity Index ($\mathbf{II}$) of our population of alternatives as:

\[ \mathbf{II} = 1 - e^{-\frac{1}{2K}\sum_{i}\left(\mathbb{D}_{KL}(\hat{S}_{i}||S^{\star}_{i}) + \mathbb{D}_{KL}(S^{\star}_{i}||\hat{S}_{i})\right)} \]

Where $\mathbb{D}_{KL}(P||Q)$ is the KL divergence of the distributions of continuous random variables $P$ and $Q$, defined as:

\[\mathbb{D}_{KL}(P||Q) = \int_{-\infty}^{\infty} p(x)log\left(\frac{p(x)}{q(x)}\right)dx \label{kl_def}\]

For Gaussian distributions $P \sim \mathcal{N}[\mu_{1},\sigma_{1}^{2}]$ and $Q \sim \mathcal{N}[\mu_{2},\sigma_{2}^{2}]$, Equation \ref{kl_def} becomes \citep{robert1996intrinsic}:

\[\mathbb{D}_{KL}(P||Q) =  log \frac{\sigma_{2}}{\sigma_{1}} + \frac{\sigma_{1}^{2} + (\mu_{1} - \mu_{2})^{2}}{2\sigma_{2}^{2}} - \frac{1}{2}\]

Using this, we can easily calculate an empirical estimate of the degree of pairwise intransitivity between alternatives. Note that $\mathbf{II} = 0$ if the predicted and observed distributions are identical, i.e. the observed sampling data corresponds precisely with the fully transitive Thurstonian model, and $\mathbf{II} = 1$ if the KL divergence between the predicted and observed distributions is infinite. Given a suitable threshold value for $\mathbf{II}$, we can define a hybrid sampling method that chooses samples according to ML-POCBAm when the currently available sampling data suggests the Thurstonian model fitted by ML-POCBAm to be plausible, and reverts to standard POCBAm for sample acquisition if the divergence between the fitted model and the observed data is too great.

\begin{figure}
\begin{subfigure}{.5\linewidth}
\centering
\includegraphics[width=\linewidth]{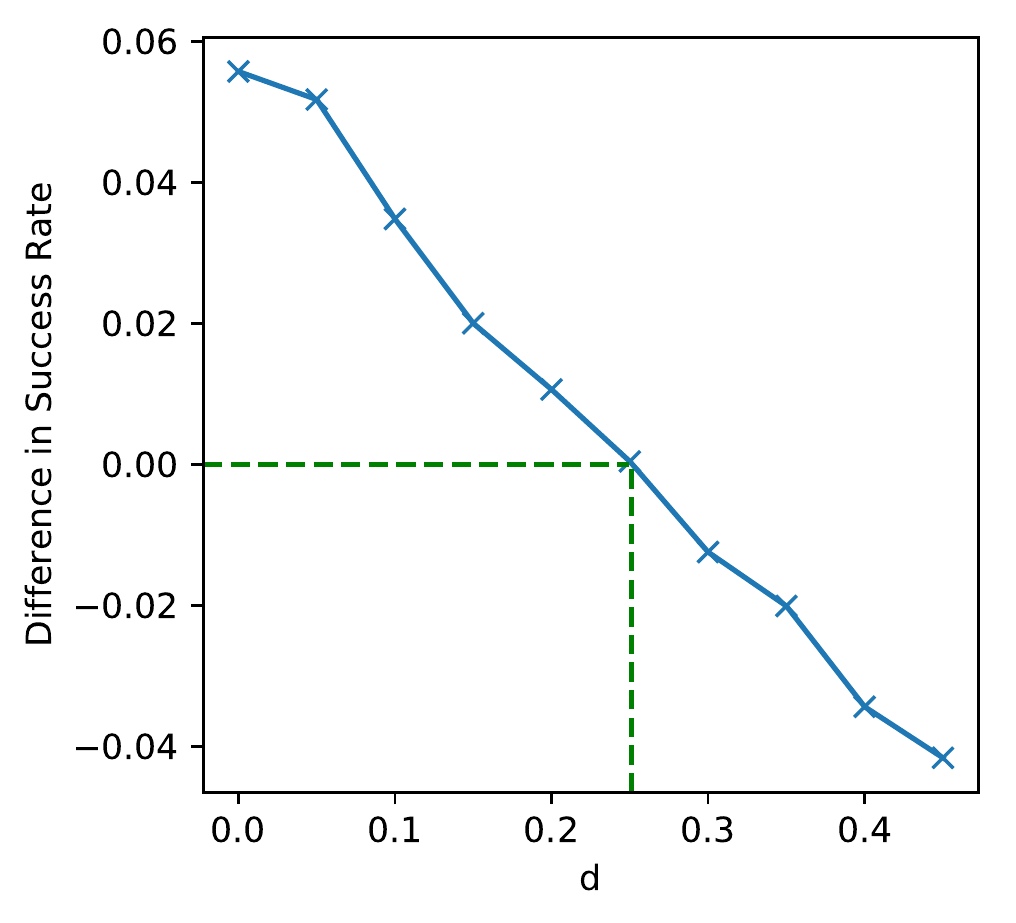}
\caption{}
\label{ii:a}
\end{subfigure}%
\begin{subfigure}{.5\linewidth}
\centering
\includegraphics[width=\linewidth]{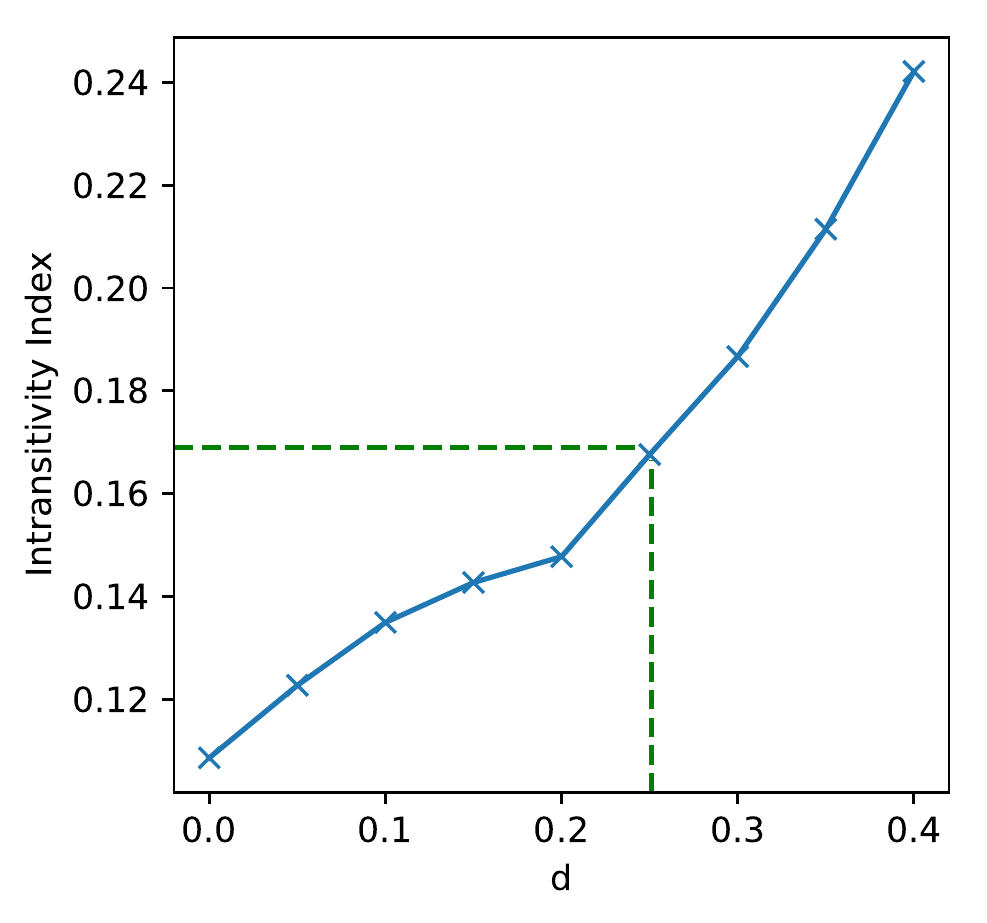}
\caption{}
\label{ii:b}
\end{subfigure}\\
\begin{subfigure}{\linewidth}
\centering
\includegraphics[width=0.5\linewidth]{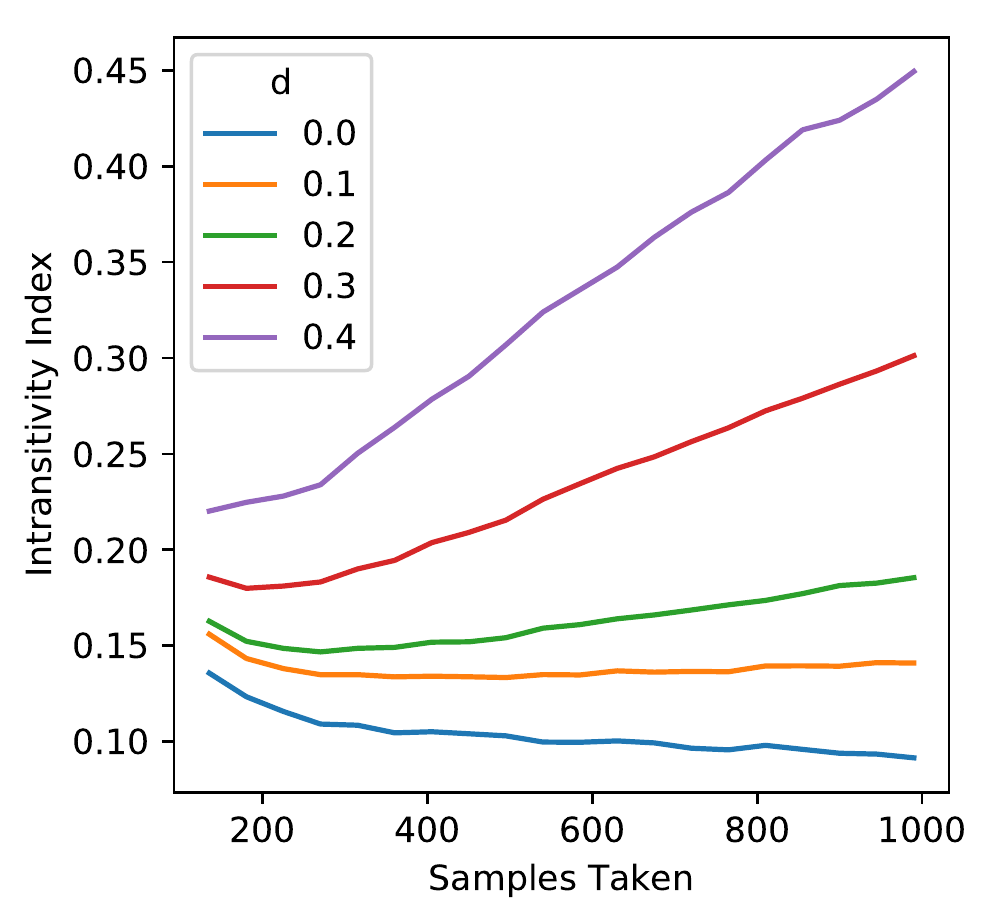}
\caption{}
\label{ii:c}
\end{subfigure}
\caption{Subfigure \ref{ii:a} shows relative performance of ML-POCBAm and POCBAm after 250 samples for different values of $d$, with estimated intransitivity index values for each different $d$ shown in \ref{ii:b}. \ref{ii:c} shows how the empirically estimated intransitivity index changes as more samples are taken for different $d$ values.}
\label{fig:ii}
\end{figure}

Figure \ref{fig:ii} shows the relationship between $d$, $\mathbf{II}$, and the difference in performance between ML-POCBAm and POCBAm after 250 samples on the noise perturbed Thurstonian model, averaged over 1000 replications. Performance for the two methods is equal at approximately $\mathbf{II} = 0.17$. In the next section we test the hybrid ML-POCBAm using this threshold value on a game player ranking problem.

%\subsection{Strongly Stochastically Transitive Value-Based preference models}
%Here we test the top-k selection performance of the LS-POCBAm method on a more general class of underlying preference model. As discussed in \cite{shah2016}, the strong assumptions imposed by parametric models like the value-based Thurstone model may be too restrictive for real-world cases where alternative fitness cannot be accurately described by the value of a single latent factor. For example in the case of ranking footballs team, where performance may depend on a wide variety factors with complex interactions (for example the individual player on the team, home advantage, weather etc.). 

\section{Top-k selection of Poker Players}\label{poker_experiments}
Recent work \citep{li2017,li2018} has considered the development of highly skilled automated Texas Hold'em playing agents, using an evolutionary strategy to optimize the weights of a recurrent neural network based player model. Typically in these evolutionary methods, a top performing portion of the population of playing agents is identified, which are then used to produce parameters for a new population of players for the next generation. Simulating games between players in order to collect the necessary information to accurately identify which players are strong and should be preserved is computationally expensive as many hands need to be played between players to overcome the stochastic effects of the card shuffle. Improving the accuracy of the top player subset or reaching the same level of accuracy with fewer samples can both be valuable.

We generate poker players using the player model described in \cite{li2018}. This represents, to the best of our knowledge, the current state-of-the-art player model for evolved no limit Texas Hold'em agents. Each player consist of three main components, a pattern recognition tree (PRT) that records the frequency of betting patterns observed during the game, along with opponent fold frequency and showdown win frequency for each betting pattern. To restrict the growth of the tree, opponent bets are discretized into seven different size buckets, and the agent's own bets are restricted to 5 different sizes (0.5x pot, 1x pot, 1.5x pot, 2x pot and all in). The data recorded in the PRT allows the agent to identify patterns in the opponent's play (like for example the tendency to fold after the agent makes a large bet) and is used to provide information for the other components of the player model: At each decision point during a hand, the agent looks up the statistics from the PRT for the betting sequence that would result from each possible action and forwards them to the other key player model components -- the opponent fold rate estimator (OFRE) network, and the showdown win rate estimator (SWRE) network.

In this experiment, we generate 10 populations of 10 candidate players, by randomly selecting weights for the OFRE and SWRE networks uniformly at random from $[-1,1]$. We generate the ground truth rankings for each population of players by playing a large number (20,000) of poker hands between each possible pair of players. Game rules are as in the AAAI Annual Computer Poker competition (http://www.computerpokercompetition.org). Player starting stacks are set to 200 Big Blinds (BB) and reset to the starting amount after each hand. The 20,000 hands between each pair of players is split into two parts, with each player playing 5000 hands from each position (playing first or second), with duplicate sets of card shuffles used in each part. Players are ranked by their cumulative win/loss amount. 

\begin{figure}
    \centering
    \includegraphics[width= \linewidth]{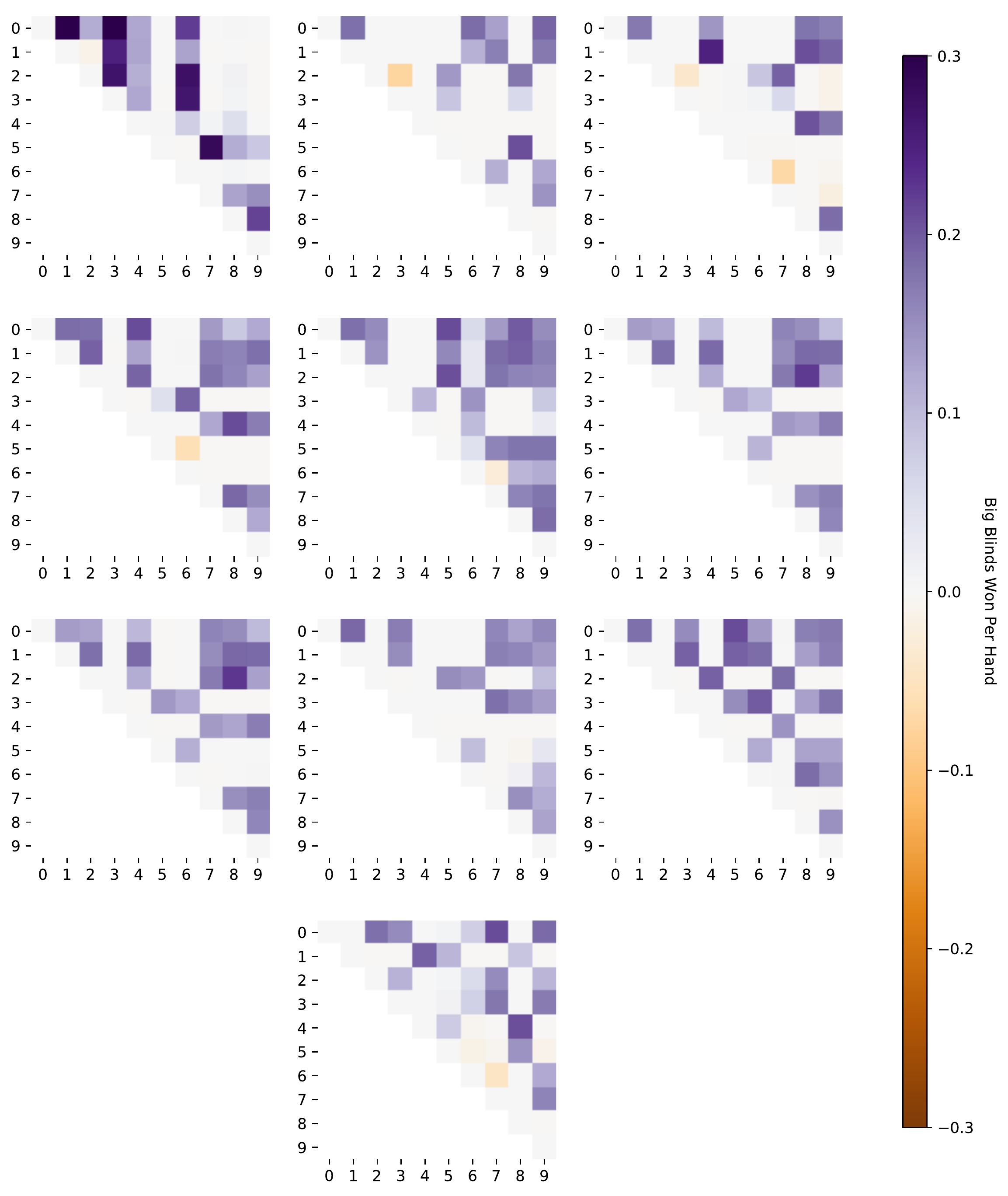}
    \caption{Visualization of the pairwise preference matrices of the ten randomly generated populations of poker players used in Section \ref{poker_experiments}. Players are indexed in true ranking order.}
    \label{fig:poker_examples}
\end{figure}

Figure \ref{fig:poker_examples} shows the pairwise performance matrices for each set of players according to these ground truth rankings. We observe that none of the pairwise rankings of the generated player groups is stochastically transitive, with negative pairwise means appearing in the upper triangle of the ordered preference matrices in all ten cases, with an average of $11.1$ of $45$ per population. However, pairwise performance of players frequently appears to be correlated, as many players who have similar (or different) performance against a given opponent also have similar (or different) performance against their other opponents, as evidenced by the clear vertical and horizontal stripes visible in the figure. As the network weights in the players were randomly generated and untrained, it was relatively common for one of the possible player actions to be dominant. The effect of this was to greatly increase the range of pairwise sampling variances. Compare for example games between two players who almost always fold immediately with games between two players who immediately raise all-in. Both pairs will have a long-term pairwise mean of 0, but the sampling variance of the latter pair will be 40,000 times greater. These large differences in variance scales, combined with the presence of intransitivity, make this a particularly challenging ranking problem.

\begin{figure}
    \centering
    \includegraphics[width= \linewidth]{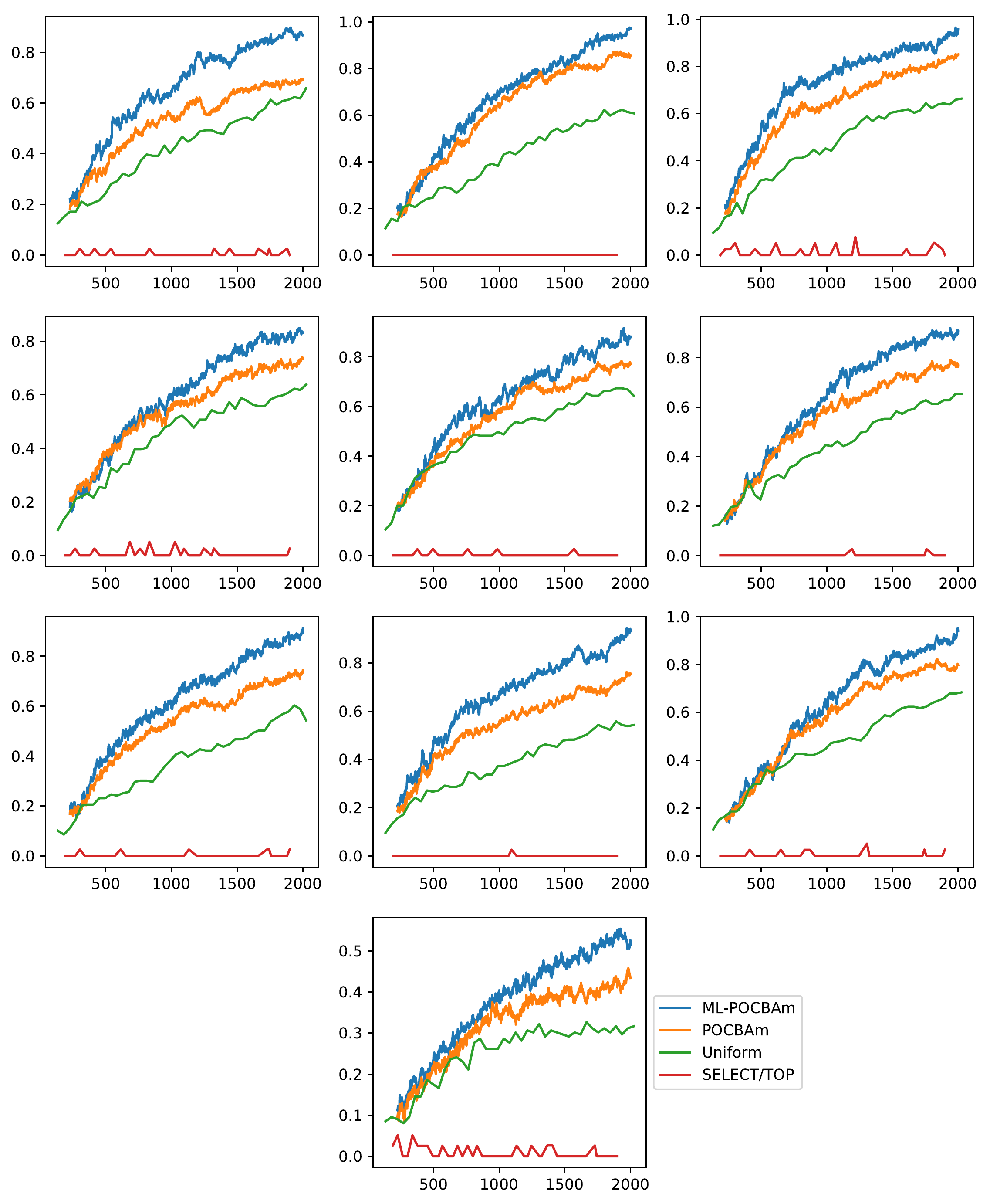}
    \caption{Performance of ML-POCBAm, POCBAm and Uniform sampling on top 4 of 10 selection for each of the populations of poker players shown in Figure \ref{fig:poker_examples}.}
    \label{fig:poker_results}
\end{figure}

Figure \ref{fig:poker_results} compares the performance of the hybrid ML-POCBAm discussed above, with $\mathbf{II}$ threshold 0.17 against standard POCBAm, SELECT/TOP and uniform sampling. Each subplot shows one of the randomly generated populations of players shown before in Figure~\ref{fig:poker_examples}. In each case, the performance of each method is averaged over 500 repetitions, with different (but common across methods) random seeds for deck shuffles in each repetition. Hybrid ML-POCBAm performs best, with similar or better performance to standard POCBAm on all 10 cases. This is perhaps unsurprising, as the hybrid method can revert to using standard POCBAm for sample acquisition wherever sample results significantly disagree with its assumed Thurstonian model. SELECT/TOP is again the worst performing method, presumably because its total ordering assumption fails to hold given the high number of pairwise intransitivities in the test populations.

\section{Summary}
In this paper, we have adapted the Thurstonian parametric model for Dueling Bandit problems with quantitative sample outcomes and proposed two sample acquisition methods that exploit this model to improve top-k selection accuracy. Both ML-POCBAm and hybrid ML-POCBAm extend and improve upon the previously published POCBAm method by selectively exploiting a parametric pairwise preference model, and significantly outperform both SELECT/TOP and uniform sampling. We suggest that this result may be useful, for example in evolutionary reinforcement learning, where each generation of the evolutionary strategy uses the top-k players to generate the generating distribution for the next generation. Improving the quality of the top-k players without increasing simulation costs can potentially lead to the learning of better final weights for the player networks in fewer generations. We intend to test this application in future work.
%\acks{}

\bibliography{transitivity_paper.bib}

\end{document}